\documentclass[10pt,twocolumn,letterpaper]{article}

\usepackage[]{cvpr}              

\definecolor{cvprblue}{rgb}{0.21,0.49,0.74}
\usepackage[pagebackref,breaklinks,colorlinks,allcolors=cvprblue]{hyperref}

\def\paperID{*****} 
\def\confName{CVPR}
\def\confYear{2025}

\title{
Detecting Mislabeled and Corrupted Data via Pointwise Mutual Information}

\author{Jinghan Yang, Jiayu Weng\\
Institute of Data Science\\
The University of Hong Kong\\
}

\begin{document}
\maketitle

\begin{figure*}[ht!]
    \centering
    \includegraphics[width=1\textwidth]{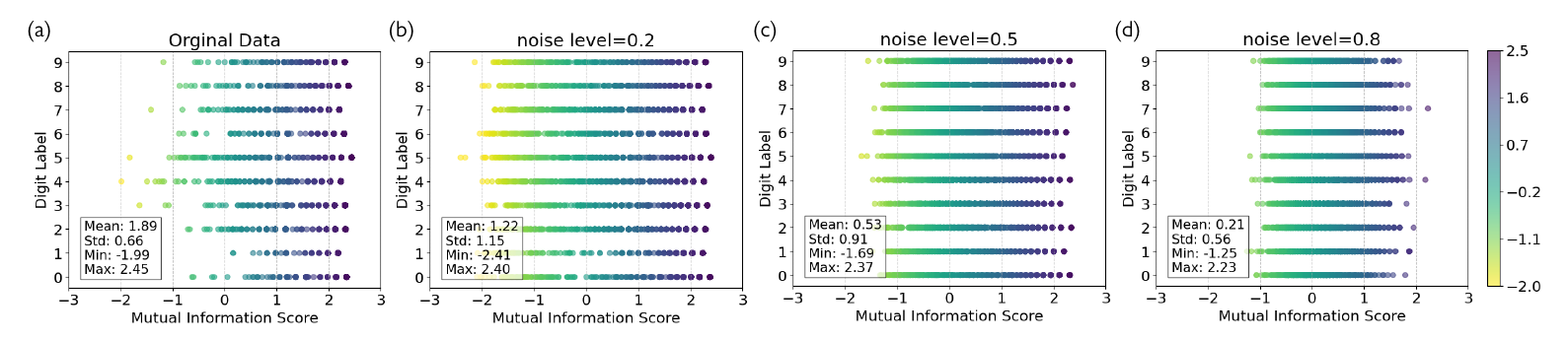}
    \caption{
    Mutual information (MI) score distributions across digit labels under increasing label noise. 
    (a) Original clean dataset.  
    (b) Corruption rate of 0.2 with randomly flipped labels.  
    (c) Corruption rate of 0.5.  
    (d) Corruption rate of 0.8.  
    Each point in the graph represents a sample and is colored according to its MI Score. }
    \label{fig:mi_label_noise}
\end{figure*}

\begin{abstract}
Deep neural networks can memorize corrupted labels, making data quality critical for model performance, yet real-world datasets are frequently compromised by both label noise and input noise. This paper proposes a mutual information-based framework for data selection under hybrid noise scenarios that quantifies statistical dependencies between inputs and labels. We compute each sample's pointwise contribution to the overall mutual information and find that lower contributions indicate noisy or mislabeled instances. 
Empirical validation on MNIST with different synthetic noise settings demonstrates that the method effectively filters low-quality samples. Under label corruption, training on high-MI samples improves classification accuracy by up to 15\% compared to random sampling. Furthermore, the method exhibits robustness to benign input modifications, preserving semantically valid data while filtering truly corrupted samples \footnote{Preprint}.
\end{abstract}


\section{Introduction}

The advancement of large-scale datasets has enabled deep neural networks to achieve remarkable success across machine learning domains including computer vision \cite{krizhevsky2012imagenet,redmon2016you}, information retrieval \cite{zhang2016deep,pang2017deeprank,onal2018neural}, and natural language processing \cite{howard2018universal,devlin2019bert,severyn2015twitter}. However, this success fundamentally depends on access to high-quality, accurately annotated data—a resource that demands significant financial and temporal investment.
In practice, annotation quality is frequently compromised. Crowdsourcing platforms and contextual labeling methods, while cost-effective, often yield inconsistent annotations \cite{paolacci2010running,mason2012conducting}. Even domain experts produce variable labels when facing complex tasks \cite{frenay2013classification,lloyd2004observer}, and adversarial actors may deliberately corrupt labels \cite{xiao2012adversarial}. These factors produce noisy labels—annotations that deviate from true class assignments. Empirical studies reveal concerning contamination rates in real-world datasets, ranging ranging from 8.0\% in ANIMAL-10N to 38.5\% in Clothing1M \cite{xiao2015learning,li2017webvision,lee2018cleannet,song2019selfie}. Such noise significantly degrades model robustness and generalization \cite{zhang2016understanding,song2022learning}, making the identification and removal of corrupted samples crucial for maintaining model performance.

Existing approaches to handle noisy labels typically fall into two categories: (1) noise-robust training methods that modify loss functions or training procedures \cite{patrini2017making,wang2019symmetric,zhang2018generalized}, and (2) sample selection methods that identify and remove corrupted instances \cite{jiang2018mentornet,han2018co,li2020dividemix}. While the former requires architectural changes and often assumes specific noise models, the latter offers a more general solution but typically relies on model-specific signals like loss values or gradients, limiting their applicability across different architectures.

We propose using Mutual Information (MI) between inputs and labels as a metric to identify noisy data points. MI quantifies statistical dependency by measuring how much knowing one variable reduces uncertainty about another \cite{per1}. Intuitively, clean samples exhibit strong statistical alignment between features and labels, while corrupted samples show weaker dependencies. By computing pointwise mutual information (PMI) for each sample—which measures how much more likely a specific input-label pair occurs together compared to random chance—we can systematically identify and exclude low-quality data. We employ the Kraskov-Stögbauer-Grassberger (KSG) estimator \cite{kraskov2004estimating, hejna2025robot} to approximate these values directly from data without requiring true probability distributions.

Our MI-based selection offers three key advantages. First, it effectively detects label noise: corrupted labels produce flattened MI score distributions with reduced inter-class variance, making unreliable samples distinguishable. Second, it maintains robustness to benign input perturbations (e.g., mild blur or occlusion) that preserve semantic content, ensuring we don't discard valuable but imperfect data. Third, it operates in a model-agnostic manner without requiring gradient or loss signals, enabling broad applicability across architectures and domains.

This work addresses two fundamental questions: (1) How does noise affect MI attribution scores, and can these scores reliably identify mislabeled samples? (2) Does MI-based selection improve model performance compared to random sampling from noisy datasets? Through experiments on MNIST with synthetic noise, we demonstrate that MI scores strongly correlate with annotation reliability and enable effective data curation even under severe label corruption.

\begin{figure*}[h!]
    \centering
    \includegraphics[width=1\textwidth]{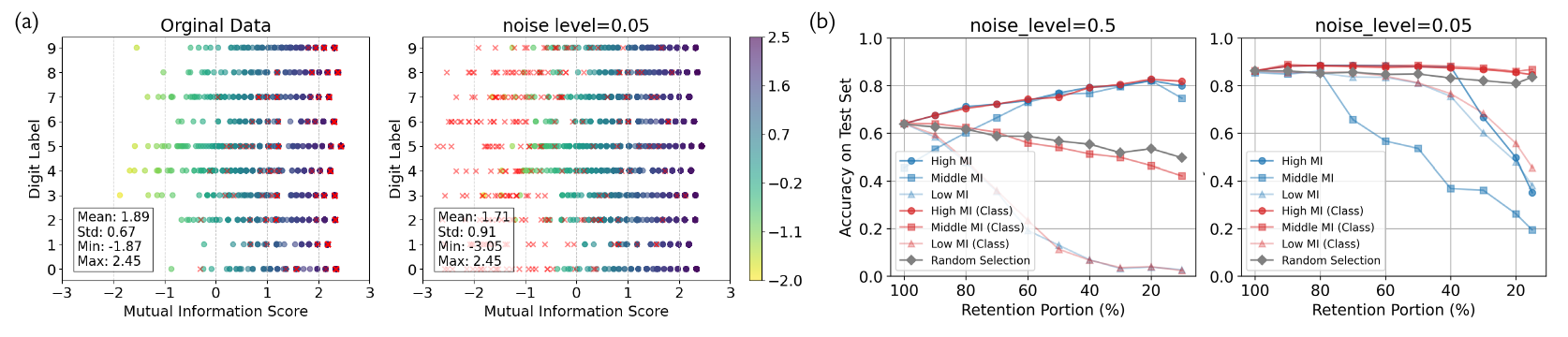}
    \caption{
    MI-based sample selection results under moderate label noise.
    (a) MI score distribution for the clean dataset (top) and the corrupted dataset with noise rate 0.05 (bottom). Red crosses indicate mislabelled samples.
    (b) Accuracy on the test set for different selection strategies as the percentage of retained training data varies. We compare selection based on global MI, class-wise MI, and random sampling.}
    \label{fig:mi_selection_strategy}
\end{figure*}

\section{Method}
\subsection{Mutual Information for Data Quality Assessment}
In supervised learning, training data quality fundamentally depends on how well inputs encode information about their labels. We propose using Mutual Information (MI) to quantify this relationship and identify problematic data points.
MI measures the statistical dependence between two variables—in our case, inputs $X$ and labels $Y$. Intuitively, it answers: "How much does knowing the input tell us about the label, and vice versa?" For clean data, we expect high MI between correctly matched input-label pairs. Conversely, mislabeled or noisy data should exhibit low MI, as the corrupted associations weaken the statistical relationship.
Formally, MI can be expressed through entropy—a measure of uncertainty in a random variable. For inputs $X$ and labels $Y$:
\begin{equation} I(X; Y) = H(Y) - H(Y|X) = H(X) - H(X|Y), \end{equation}
where $H(Y)$ is the entropy of labels and $H(Y|X)$ is the conditional entropy of labels given inputs. This formulation reveals MI as the reduction in uncertainty about one variable when the other is known.
For practical computation, MI can be written as: \begin{equation} I(X; Y) = \sum_{x \in \mathcal{X}} \sum_{y \in \mathcal{Y}} P(x, y) \log \left( \frac{P(x, y)}{P(x)P(y)} \right), \end{equation}
where the term $\log \frac{P(x,y)}{P(x)P(y)}$ represents the pointwise mutual information (PMI) for each $(x,y)$ pair.
While global MI characterizes the overall dataset quality, identifying specific problematic samples requires decomposing this global measure into local contributions. Each data point $(x_i, y_i)$ contributes differently to the total MI—clean samples reinforce the statistical dependence while noisy samples weaken it.
This motivates our key insight: by quantifying each point's contribution to the global MI, we can identify and remove samples that degrade data quality. However, this requires:
Practical estimation: Real-world datasets lack known probability distributions
Local decomposition: Attribution of MI to individual data points
\subsection{The KSG Estimator: A Practical Solution}
The Kraskov-Stögbauer-Grassberger (KSG) estimator \cite{kraskov2004estimating} elegantly addresses both challenges. It provides a non-parametric method to estimate MI directly from data while naturally decomposing it into local contributions.
For a dataset of $N$ input-label pairs ${(x_i, y_i)}_{i=1}^N$, KSG defines the local MI contribution of each point as:
\begin{equation} I(x_i; y_i) = \psi(k) + \psi(N) - \psi(n_x(i) + 1) - \psi(n_y(i) + 1), \end{equation}
where:
$\psi(\cdot)$ is the digamma function
$k$ is the number of nearest neighbors (hyperparameter)
$n_x(i)$ and $n_y(i)$ count neighbors in the marginal spaces
The global MI equals the average of local contributions: $I(X; Y) = \frac{1}{N} \sum_{i=1}^{N} I(x_i; y_i)$

The local MI contribution $I(x_i; y_i)$ has a clear geometric interpretation that directly informs data quality:
High $I(x_i; y_i)$: The point has many neighbors in the joint $(X,Y)$ space but few in the marginal spaces. This indicates strong local coupling—the input strongly predicts its label. Example: a clear “7” image surrounded by other correctly labeled “7”s.
Low/negative $I(x_i; y_i)$: The point has similar neighbor densities in joint and marginal spaces, indicating weak coupling. This suggests the input provides little information about its label. Example: a mislabeled or corrupted image whose features appear across multiple classes.
By ranking points by their local MI contributions and removing those with the lowest scores, we systematically eliminate samples that fail to encode meaningful input-label relationships, thereby improving overall data quality.

\begin{figure*}[h!]
    \centering
    \includegraphics[width=1\textwidth]{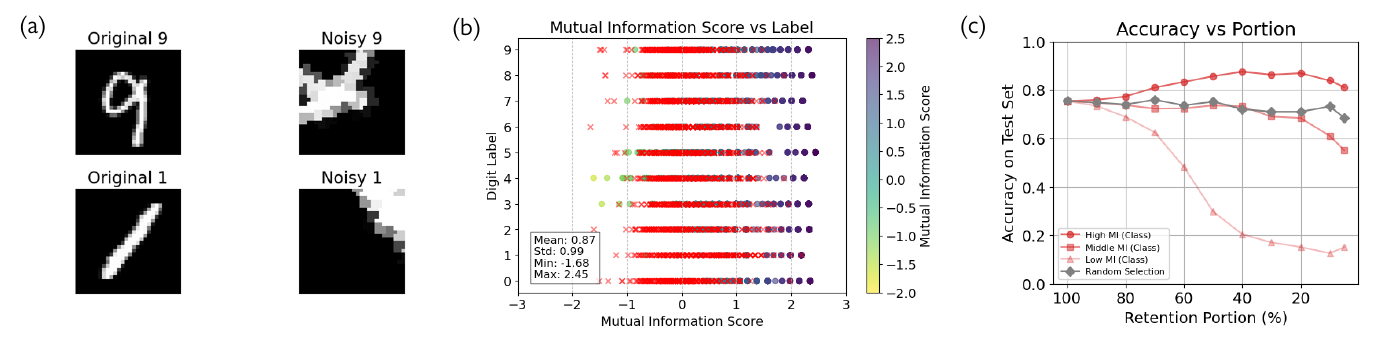}
    \caption{
Evaluation under strong geometric distortions (random affine transformations).
(a) Example pairs of clean and warped digit images with significant structural deformation.
(b) Mutual information scores plotted by class label.
(c) Test accuracy as a function of retained training data, comparing MI-based selection and random sampling strategies.
}
    \label{fig:mi_warping}
\end{figure*}

\section{Experiment}

To validate the effectiveness of mutual information as a principled metric for identifying and filtering noisy training data, we design controlled experiments on the MNIST dataset with systematically introduced corruptions.

\subsection{Experiment Setup}

\paragraph{Dataset}
We conduct experiments on the MNIST dataset, a benchmark for handwritten digit classification containing 60,000 training and 10,000 test grayscale images (28×28 pixels) across 10 digit classes (0–9). While MNIST is inherently clean, we introduce controlled synthetic noise to evaluate MI-based selection under realistic data quality conditions. 

\vspace{-10pt}
\paragraph{Model and MI Computation}
We use logistic regression as our classifier for its simplicity and interpretability. 
To address the high dimensionality of raw images for MI estimation, we employ a Variational Autoencoder (VAE) to compress the data into a lower-dimensional latent space ~\cite{hejna2025robot}.
Mutual information scores are computed in the VAE latent space between input representations and their corresponding labels. 

\vspace{-10pt}
\paragraph{Noise Injection}
To systematically evaluate MI's discriminative capabilities, we introduce two types of synthetic corruption:
\begin{itemize}
    \item \textbf{Label noise}: Training labels are randomly flipped to incorrect classes at rates of 0.2, 0.5, and 0.8, simulating increasing levels of annotation errors.
    \item \textbf{Input modifications}: Images undergo three distinct transformation categories:
    \begin{itemize}
        \item Strong geometric transformations that significantly distort digit shapes, compromising semantic identity
        \item Additive Gaussian noise that introduces visual interference while preserving underlying digit structure  
        \item Mild augmentations (light warping, small-angle rotations, moderate scaling) that simulate natural variations
    \end{itemize}
\end{itemize}

\subsection{Results}

\noindent{\textbf{Lower MI values correspond to noisy points in the distribution.}}

We first investigate how mutual information (MI) scores respond to increasing label noise. Figure~\ref{fig:mi_label_noise} shows this relationship across different noise levels. Panel (a) displays the original MI distribution without noise. We then introduce label noise by randomly flipping labels at rates of 0.2, 0.5, and 0.8.
As the noise level increases from 0.2 to 0.5, the MI score of the distribution decreases dramatically from 1.22 to 0.53, indicating weakened statistical dependency between inputs and their corrupted labels. More importantly, we observe that mislabeled points consistently exhibit negative MI values in ~\ref{fig:mi_selection_strategy}, effectively separating them from correctly labeled samples. This clear separation enables our method to both quantify the overall noise level in a dataset and identify individual noisy points, providing a foundation for filtering corrupted samples during model training in subsequent sections.

\vspace{0.5em}
\noindent{\textbf{Filtering harmful noise improves model performance.}

We evaluate several data selection strategies for model training, as illustrated in Figure~\ref{fig:mi_selection_strategy}(b). These strategies include selecting the top, middle, or bottom-ranked samples by MI score, applied either globally across the entire dataset or within each class.
As expected, selecting high-MI samples leads to improved test accuracy, since these data points exhibit stronger mutual dependency between images and labels, indicating they are more likely to be correctly annotated and semantically aligned. Conversely, training on low-MI samples results in sharp accuracy drops, confirming their detrimental impact on classification.
Notably, class-wise selection—ranking samples by MI within each class—yields more stable performance across varying retention ratios, as it preserves class balance during selection. This advantage becomes particularly important in low-data regimes or when dealing with imbalanced datasets.
When noise levels are relatively low, models trained on the full dataset already achieve high accuracy, and our MI-based filtering maintains this performance even when reducing the training set size. Under higher corruption levels, however, MI-based selection significantly outperforms random selection in both accuracy and stability, demonstrating its robustness in identifying trustworthy training examples.

\vspace{0.5em}
\noindent{\textbf{MI distinguishes semantic-altering modifications from benign variations.}}

While previous experiments focus on label noise, understanding how MI responds to different types of input modifications is also important. In real-world settings, images naturally exhibit variations due to collection conditions, compression artifacts, or environmental factors. We therefore investigate whether MI can differentiate between modifications that alter semantic content versus those that preserve it.

We examine three distinct categories of input modifications: (1) strong geometric transformations that fundamentally alter digit identity, (2) additive Gaussian noise that overlays visual interference, and (3) mild augmentations including scaling and rotation that preserve digit structure. These transformations are applied to randomly selected training samples while keeping labels unchanged.

Our results reveal a clear dichotomy in MI's response. Strong geometric transformations (Fig.~\ref{fig:mi_warping}) that distort digits beyond recognition consistently produce low MI scores, demonstrating that MI correctly identifies when semantic information has been compromised. These transformations effectively break the meaningful connection between the visual input and its label, which MI captures through drastically reduced scores.

Conversely, modifications that preserve semantic content yield markedly different results. Images with Gaussian noise (Fig.~\ref{fig:mi_gaussian}) or mild augmentations (Fig.~\ref{fig:mi_light})—despite visible changes to appearance—maintain stable MI scores. This stability occurs because these modifications, while altering surface-level features, leave the core digit identity intact. The preserved semantic structure ensures that the mutual information between input and label remains high.

This discriminative behavior addresses a critical concern: could MI-based selection inadvertently filter out useful training variations? Modern training pipelines rely heavily on data augmentation for improved generalization. Our findings demonstrate that MI exhibits precisely the selectivity needed—it flags inputs where semantic meaning is genuinely compromised while accepting those with benign variations. This makes MI-based selection not only compatible with but complementary to standard augmentation strategies, as it filters truly problematic samples while preserving beneficial diversity in the training data.

\begin{figure*}[h!]
    \centering
    \includegraphics[width=1\textwidth]{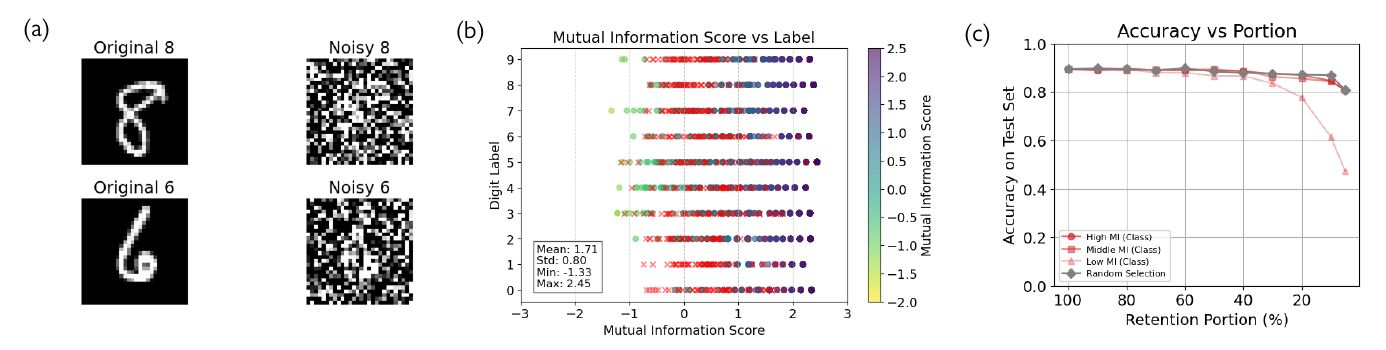}
    \caption{
Evaluation under additive Gaussian noise (noise factor = 0.9).
(a) Pairs of clean and noisy digit images, with semantic content preserved despite high pixel-level corruption.
(b) MI score distribution across labels.
(c) Accuracy curves for various data selection methods across different retention levels.
}
    \label{fig:mi_gaussian}
\end{figure*}

\begin{figure*}[h!]
    \centering
    \includegraphics[width=\textwidth]{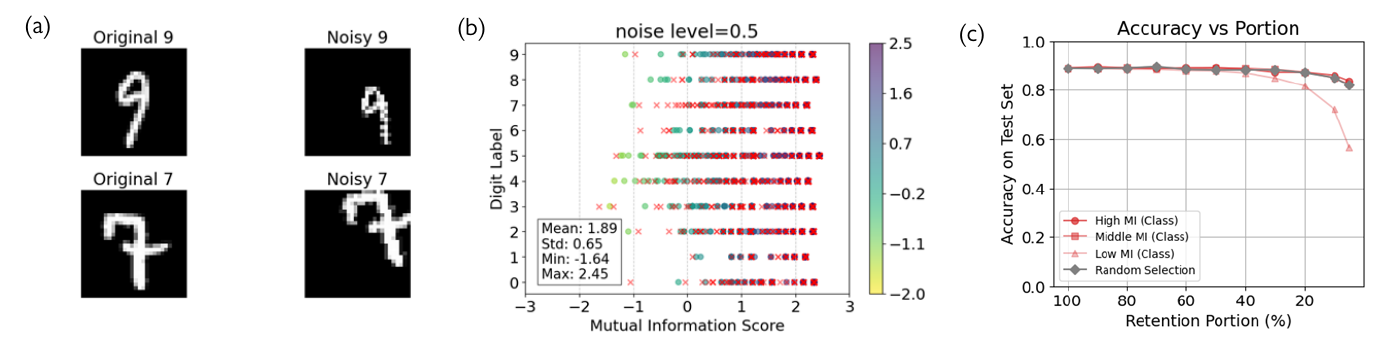}
    \caption{
    Evaluation under mild geometric transformations (e.g., rotation, scaling, and affine distortion).  
    (a) Representative examples of original versus transformed digit images.  
    (b) MI score distribution across digit classes after applying input-level transformations. Red crosses denote samples with corrupted labels.  
    (c) Test accuracy under different sample selection strategies as a function of retention ratio.
    }
    \label{fig:mi_light}
\end{figure*}

\section{Related Work}

Data quality remains a critical challenge in machine learning, particularly when training datasets suffer from label noise (randomly flipped labels) or input noise (e.g., Gaussian perturbations, occlusions, affine warping). Label noise disrupts the semantic alignment between inputs and targets, while input noise degrades feature integrity through transformations like blurring or geometric distortions. Real-world datasets exhibit concerning contamination rates, ranging from 8.0\% in ANIMAL-10N to 38.5\% in Clothing1M \cite{xiao2015learning, li2017webvision, lee2018cleannet, song2019selfie}. These imperfections degrade model performance and generalization, as DNNs can memorize even completely random labels \cite{zhang2016understanding}, necessitating robust strategies for identifying reliable training samples \cite{frenay2013classification, zhu2004class}.

Prior work has explored three dominant paradigms for handling noisy labels: loss correction, sample selection, and regularization. Loss correction techniques such as forward/backward loss correction \cite{patrini2017making}, Gold loss correction \cite{hendrycks2018using}, and symmetric cross-entropy \cite{wang2019symmetric} modify loss functions to account for noise transition matrices. Recent advances like T-Revision \cite{xia2019anchor} operate without anchor points, while Dual-T \cite{yao2020dual} factorizes the transition matrix for improved estimation. However, these require accurate estimation of noise patterns, which is often infeasible in real-world scenarios with complex or instance-dependent noise \cite{chen2021beyond}. Sample selection methods like MentorNet \cite{jiang2018mentornet}, Co-teaching \cite{han2018co}, and DivideMix \cite{li2020dividemix} use auxiliary networks or small-loss criteria to filter noisy samples during training, exploiting the memorization effect where DNNs learn clean patterns before fitting noise \cite{arpit2017closer}. SELFIE \cite{song2019selfie} identifies refurbishable samples through prediction consistency. Influence functions identify impactful training points by estimating how model predictions change with sample modification \cite{koh2017understanding} and can be applied to noisy data filtering \cite{teso2021interactive}. It has been found that to change a prediction, the number of training data points to relabel is related to the noise ratio in the training set \cite{yang2023relabeling}. While effective, they rely on model gradients or loss dynamics, limiting their applicability to specific architectures and optimization settings. Regularization approaches such as Mixup \cite{zhang2017mixup}, label smoothing \cite{lukasik2020does, wei2021smooth}, and robust early-learning \cite{xia2021robust} impose inductive biases to reduce overfitting to noisy labels. Pre-training provides robust initialization \cite{hendrycks2019using}, while early stopping prevents noise memorization \cite{liu2020early}. However, these techniques may inadvertently suppress discriminative features in clean data and often fail under severe label corruption \cite{song2022learning}. Our work introduces a complementary paradigm: MI-based data filtering, which bypasses reliance on loss dynamics, noise transition matrices, or architectural priors. Unlike gradient-dependent methods, our approach operates directly on the input-label dependency structure, enabling model-agnostic noise detection.

Mutual information has proven effective for data selection across diverse machine learning contexts. \citet{mak2024statistical} propose MI-based stratified sampling combined with support points to address class imbalance, achieving over 80\% data reduction while preserving classification accuracy through representative subset selection. Similarly, \citet{kothawade2021submodular} leverage submodular mutual information functions for targeted data subset selection, demonstrating 20-30\% accuracy gains when selecting samples that align with specific target distributions. In the context of Gaussian Processes, \citet{zainudin2012mutual} employ MI to iteratively select informative data points for people tracking, using the Mahalanobis distance to determine when new observations provide additional information beyond the current model. Recent work has also explored MI for robust loss functions \cite{ghosh2017robust, zhang2018generalized} and meta-learning approaches \cite{shu2019meta}, though these require specific architectural modifications or clean validation data. These works establish MI as a principled criterion for data selection, focusing primarily on computational efficiency, class balance, or domain-specific applications. 
\citet{hejna2025robot} utilize mutual information to prioritize trajectories with higher dependency between states and actions, thereby improving performance in robot training tasks.
Our work extends this line of research by demonstrating that MI-based selection can effectively distinguish between clean and noisy samples, providing a unified framework for data quality assessment that complements existing selection strategies.

Our work distinguishes itself through three distinct contributions. First, we adapt the KSG estimator \cite{kraskov2004estimating, hejna2025robot} to compute pointwise MI contributions (Section 3.2) for fine-grained identification of noisy samples without requiring labeled validation data or model retraining. Second, we demonstrate MI's robustness to hybrid noise scenarios (simultaneous label and input noise), a critical advantage over methods addressing either noise type in isolation \cite{jiang2018mentornet, han2018co, zhang2017mixup, arazo2019unsupervised}. Third, unlike existing approaches that require noise rate estimation \cite{han2018co, song2019selfie} or transition matrix computation \cite{patrini2017making, yao2020dual}, our method operates without prior knowledge of noise characteristics, making it practical for real-world deployment where such information is unavailable.

\section{Conclusion}

In this work, we propose a data selection method based on mutual information (MI) between inputs and labels of data. The method attribute MI scores to individual data points, and filter out samples with low MI attribution relative to the distribution.
Empirical results show that lower MI scores effectively identify corrupted or mislabeled samples, distinguishing them from clean, semantically aligned data. In noisy settings, selecting high-MI-attribution points improves model accuracy compared to random sampling.
In addition, we find that MI remains robust to benign visual noise added to input images , which does not alter their semantic meaning, and specifically flags truly uninformative or misleading inputs.

As a first step, our current study is limited to logistic regression models and MNIST dataset.
Future work will explore the extension of this framework to more complex models and benchmarks such as CIFAR-10 and TinyImageNet. We will also include comparisons with stronger baselines.

{
    \small
    \bibliographystyle{ieeenat_fullname}
    \bibliography{main}
}


\end{document}